\pdfoutput=1

\documentclass[11pt]{article}

\usepackage{acl}

\usepackage{times}
\usepackage{latexsym}
\usepackage{graphicx}

\usepackage[T1]{fontenc}

\usepackage[utf8]{inputenc}

\usepackage{microtype}

\usepackage{inconsolata}

\title{MasonTigers at SemEval-2024 Task 1: An Ensemble Approach for Semantic Textual Relatedness}

 \author{Dhiman Goswami, Sadiya Sayara Chowdhury Puspo, Md Nishat Raihan, \\ {\bf Al Nahian Bin Emran, Amrita Ganguly, Marcos Zampieri } \\ George Mason University, USA \\
 \texttt{dgoswam@gmu.edu}
 }
 
\begin{document}
\maketitle
\begin{abstract}

This paper presents the \textit{MasonTigers}' entry to the SemEval-2024 Task 1 - Semantic Textual Relatedness. The task encompasses supervised (Track A), unsupervised (Track B), and cross-lingual (Track C) approaches to semantic textual relatedness across 14 languages. \textit{MasonTigers} stands out as one of the two teams who participated in all languages across the three tracks. Our approaches achieved rankings ranging from $11^{th}$ to $21^{st}$ in Track A, from $1^{st}$ to $8^{th}$ in Track B, and from $5^{th}$ to $12^{th}$ in Track C. Adhering to the task-specific constraints, our best performing approaches utilize an ensemble of statistical machine learning approaches combined with language-specific BERT based models and sentence transformers.

\end{abstract}

\section{Introduction}

In this modern era of information retrieval and NLP, understanding semantic relatedness is fundamental for refining and optimizing diverse applications. Semantic relatedness refers to the degree of similarity and cohesion \cite{hasan1976cohesion} in meaning between two words, phrases, or sentences. Semantic relatedness allows systems to grasp the contextual and conceptual connections between words or expressions. Various NLP tasks and applications can benefit from modeling semantic relatedness such as question answering \cite{park2014isoft}, knowledge transfer \cite{rohrbach2010helps},  text summarization \cite{rahman2023query}, machine translation \cite{ali2009improving}, and content recommendation \cite{piao2016computing}. 

While significant research has been conducted on semantic relatedness in English, more recently the interest in semantic relatedness in other languages has been steadily growing \cite{baldissin2022diawug}. This reflects an increasing awareness of the need for developing models to languages English other than English. NLP is evolving rapidly and we have been witnessing the emergence of language-specific transformer, the release of datasets for downstream tasks in diverse languages, and the development of multilingual models designed to handle linguistic diversity. 

SemEval-2024 Task 1 - Semantic Textual Relatedness \cite{semrel2024task} aims to determine the semantic textual relatedness (STR) of sentence pairs across 14 diverse languages. Track A focuses on nine languages (Algerian Arabic, Amharic, English, Hausa, Kinyarwanda, Marathi, Moroccan Arabic, Spanish, Telugu) using a supervised approach where systems are trained on labeled training datasets. Track B adopts an unsupervised approach, prohibiting the use of labeled data to indicate similarity between text units exceeding two words.  
This track encompasses 12 languages (Afrikaans, Algerian Arabic, Amharic, English, Hausa, Hindi, Indonesian, Kinyarwanda, Modern Standard Arabic, Moroccan Arabic Punjabi, and Spanish). Track C involves cross-lingual analysis across the 12 aforementioned languages. 
Participants in this track must utilize labeled training data from another track for at least one language, excluding the target language. Evaluation across all three tracks involves using Spearman Correlation between predicted similarity scores and human-annotated gold scores. We conduct distinct experiments for each track using statistical machine learning approaches along with the embeddings generated by transformer based models.


\section{Related Work}

Understanding the level of semantic relatedness between two languages has been regarded as essential for grasping their meaning. Notable studies on the topic including  \citet{agirre2012semeval, agirre2013sem, agirre2014semeval, agirre2015semeval, agirre2016semeval,dolan2005automatically} and \citet{li2006sentence} have introduced datasets like STS, MRPC, and LiSent. These datasets have been pivotal in advancing research in tasks such as text summarization and plagiarism detection. 


Finding semantic relatedness and semantic similarity, as well as determining sentence pair similarity using existing datasets or paired annotation, are integral in understanding the nuances of language comprehension. Previous studies describe how words and sentences are perceived to convey similar meanings \cite{halliday2014cohesion,morris1991lexical,asaadi2019big,abdalla2021makes,goswami2024masontigers}.

Methodologies like paired comparison represent the most straightforward type of comparative annotations \cite{thurstone1994law}, \cite{david1963method}. Best-Worst Scaling (BWS) \cite{louviere1991best}  a comparative annotation schema, offer insights into methods for evaluating relatedness through pairwise comparisons. The utilization of these methods aids in generating ordinal rankings of items based on their semantic relatedness. \citet{kiritchenko2016capturing, kiritchenko2017best} highlight the effectiveness of such techniques, emphasizing the importance of reliable scoring mechanisms derived from comparative annotations for understanding the intricacies of semantic relatedness in natural language processing tasks.

\section{Data}

The shared task comprises three tracks: Supervised, Unsupervised, and Cross-Lingual. 
The dataset \cite{ousidhoum2024semrel2024} is comprised of two columns: the initial column, labeled "text," containing two full sentences separated by a special character, and the second column, labeled as "score", which includes degree of semantic textual relatedness for the corresponding pair of sentences. In the supervised track (Track A), there are 9 languages, and for each language, train, dev, and test sets are provided. The specifics of the dataset for this track can be found in Table \ref{tab:trackA DS}.

\begin{table}[!h]
\centering
\scalebox{.90}{
\begin{tabular}{lccc}
\hline
\textbf{Language} & \textbf{Train} & \textbf{Dev} & \textbf{Test} \\
\hline
Algerian Arabic (arq) & 1,261 & 97 & 583 \\
Amharic (amh) & 992 & 95 & 171 \\
English (eng) & 5,500 & 250 & 2,600 \\
Hausa (hau) & 1,736 & 212 & 603 \\
Kinyarwanda (kin) & 778 & 102 & 222 \\
Marathi (mar) & 1,200 & 293 & 298 \\
Moroccan Arabic (ary) & 924 & 71 & 426 \\
Spanish (esp) & 1,562 & 140 & 600 \\
Telugu (tel) & 1,170 & 130 & 297 \\
\hline
\end{tabular}
}
\caption{Track A Dataset Distribution}
\label{tab:trackA DS}
\end{table}

\noindent In the unsupervised track (Track B), there are 12 languages and for all the languages dev and test set is provided. The details of the dataset of this track is available in Table \ref{tab:trackB DS}.

\begin{table}[!h]
\centering
\scalebox{.90}{
\begin{tabular}{lcc}
\hline
\textbf{Language} & \textbf{Dev} & \textbf{Test} \\
\hline
Afrikaans (afr) & 20 & 375 \\
Algerian Arabic (arq) & 97 & 583 \\
Amharic (amh) & 95 & 171 \\
English (eng) & 250 & 2,600 \\
Hausa (hau) & 212 & 603 \\
Hindi (hin) & 288 & 968 \\
Indonesian (ind) & 144 & 360  \\
Kinyarwanda (kin) & 102 & 222 \\
Modern Standard Arabic (arb) & 32 & 595 \\
Moroccan Arabic (ary) & 71 & 426 \\
Punjabi (pan) & 242 & 634 \\
Spanish (esp) & 140 & 600 \\
\hline
\end{tabular}
}
\caption{Track B Dataset Distribution}
\label{tab:trackB DS}
\end{table}

\noindent Finally, in the cross-lingual track (Track C), there are 12 languages and for all the languages dev and test set is provided and they are same as the unsupervised track. Here the training dataset is not provided. Hence, for each individual language of this track, we select 5 languages from supervised track (different from the target language) and merge training data of those five languages to create the training dataset for each of the languages of cross-lingual track. The details of the dataset of this track is available in Table \ref{tab:language_split}.

\begin{table*}[!h]
\centering
 \scalebox{.90}{
\begin{tabular}{lcccc}
\hline
\textbf{Language} & \textbf{Train Data from (Track A)} & \textbf{Train} & \textbf{Dev} & \textbf{Test} \\
\hline
Afrikaans (afr) & amh, eng, esp, arq, ary & 10,239 & 20 & 375\\
Algerian Arabic (arq) & amh, hau, esp, eng, ary & 10,714 & 97 & 583\\
Amharic (amh) & eng, hau, esp, arq, ary & 10,983 & 95 & 171\\
English (eng) & arq, ary, mar, esp, tel & 6,117 & 250 & 2,600\\
Hausa (hau) & amh, esp, arq, ary, eng & 10,239 & 212 & 603\\
Hindi (hin) & esp, eng, mar, ary, tel & 10,356 & 288 & 968\\
Indonesian (ind) & ary, eng, mar, esp, tel & 5,356 & 144 & 360\\
Kinyarwanda (kin) & amh, esp, ary, arq, eng & 10,239 & 102 & 222\\
Modern Standard Arabic (arb) & amh, eng, arq, esp, ary & 10,239 & 32 & 595\\
Moroccan Arabic (ary) & amh, hau, eng, esp, arq & 11,051 & 71 & 426\\
Punjabi (pan) & arq, esp, mar, eng, tel & 10,693 & 242 & 634\\
Spanish (esp) & arq, ary, mar, eng, tel & 10,055 & 140 & 600\\
\hline
\end{tabular}
}
\caption{Track C Data Distribution (Train Data from Track A)}
\label{tab:language_split}
\end{table*}

\section{Experiments}

We use statistical machine learning along with language specific BERT-based models to find the sentence embeddings and predict relatedness between pair of sentences. Additionally, we use sentence transformers for the supervised track. Our experiments are described in detail in the next sections.

\subsection{Track A - Supervised}

At first, we find the embedding of the training data using Term Frequency - Inverse Document Frequency (TF-IDF) \cite{aizawa2003information}, Positive Point-wise Mutual Information (PPMI) \cite{church1990word}, and Language-Agnostic BERT Sentence Embedding (LaBSE sentence transformer) \cite{feng2020language} separately. Also we find the embeddings using language specific BERT based models. For Algerian Arabic, Amharic, English, Hausa, Kinyarwanda, Marathi, Moroccan Arabic, Spanish and Telugu - DziriBERT \cite{abdaoui2021dziribert}, AmRoBERTa \cite{yimam2021introducing}, RoBERTa \cite{liu2019roberta}, HauRoBERTa \cite{adelani2022masakhaner}, KinyaBERT \cite{nzeyimana2022kinyabert}, MarathiBERT \cite{joshi2022}, DarijaBERT \cite{gaanoun2024darijabert}, SpanishBERT \cite{CaneteCFP2020} and TeluguBERT \cite{joshi2022l3cubehind} are used.

For each training embedding, we calculate the cosine similarity \cite{rahutomo2012semantic} between the pairs. After that we apply ElasticNet \cite{zou2005regularization} and Linear Regression \cite{gross2003linear} separately on these embeddings and predict the relatedness of the sentence pairs in the development phase. We clip the predicted values to ensure the prediction range from 0 to 1. In the development phase, we  find the Spearman Correlation Coefficient \cite{myers2004spearman} of these eight predictions (four each by ElasticNet and Linear Regression). Finally, we perform a weighted ensemble depending on the Spearman Correlation Coefficient of the eight predicted results and get our ensembled Spearman Correlation Coefficient in development phase. We also perform this approach on the test data and find our best Spearman Correlation Coefficient in the evaluation phase.

\subsection{Track B - Unsupervised}

For unsupervised track, we find the embedding of the development data using Term Frequency - Inverse Document Frequency (TF-IDF) \cite{aizawa2003information} and Positive Point-wise Mutual Information (PPMI) \cite{church1990word} separately. Also we find the embeddings using language specific BERT based models. For Afrikaans, Algerian Arabic, Amharic, English, Hausa, Hindi, Indonesian, Kinyarwanda, Modern Standard Arabic, Moroccan Arabic, Punjabi and Spanish - AfricanBERTa, DziriBERT \cite{abdaoui2021dziribert}, AmRoBERTa \cite{yimam2021introducing}, RoBERTa \cite{liu2019roberta}, HauRoBERTa \cite{adelani2022masakhaner}, HindiBERT \cite{joshi2022l3cubehind}, IndoBERT \cite{koto2020indolem}, KinyaBERT \cite{nzeyimana2022kinyabert}, ArabicBERT \cite{safaya2020kuisail}, DarijaBERT \cite{gaanoun2024darijabert}, PunjabiBERT \cite{joshi2022l3cubehind}, and SpanishBERT \cite{CaneteCFP2020} are used.

Then for each development embedding generated by these three approaches, we calculate cosine similarity \cite{rahutomo2012semantic} between the pairs. In the development phase, we find the Spearman correlation \cite{myers2004spearman} of these values calculated on embeddings found by three different procedures and perform an average ensemble of the calculated results to get our ensembled Spearman correlation in development phase. We also perform this approach on the test data and find our best Spearman correlation in the evaluation phase.

\subsection{Track C - Cross-Lingual}

For each language in cross-lingual track, we select 5 different languages from Supervised Track to use as training data. The details of the language selection is provided in Table \ref{tab:language_split}. The we find the embedding of the training data using Term Frequency - Inverse Document Frequency (TF-IDF) \cite{aizawa2003information} and Positive Point-wise Mutual Information (PPMI) \cite{church1990word} separately. Also we find the embeddings using language specific (unrelated to the target language) BERT based models. For Afrikaans, Amharic, Hausa and Kinyarwanda - we use ArabicBERT \cite{safaya2020kuisail}, for Algerian Arabic, Modern Standard Arabic and Moroccan Arabic - we use AfricanBERTa\footnote{\url{https://huggingface.co/mrm8488/AfricanBERTa}}, for English, Hindi, Indonesian, Punjabi and Spanish - SpanishBERT \cite{CaneteCFP2020}, BanglaBERT \cite{bhattacharjee2022banglabert}, RoBERTa-tagalog \cite{cruz2021improving}, HindiBERT \cite{joshi2022l3cubehind} and RoBERTa \cite{liu2019roberta} are used. Then for each training embedding generated by these three approaches, we calculate cosine similarity \cite{rahutomo2012semantic} between the pairs. After that we apply ElasticNet \cite{zou2005regularization} and Linear Regression \cite{gross2003linear} separately on these embeddings and predict the similarity of the sentence pairs in the development phase. We clip the predicted values to ensure the prediction range from 0 to 1. In the development phase, we  find the Spearman correlation of these six predictions (three each by ElasticNet and Linear Regression) and perform an average ensemble of the predictions to get our ensembled Spearman correlation in development phase. We also perform this approach on the test data and find our best Spearman correlation in the evaluation phase.

\section{Results}

For all the tracks, ensemble of the predictions prove helpful in terms of achieving better Spearman correlation.

For Track A sentence transformer LaBSE along with Linear Regression performs the best among the eight combinations for all the languages. Then the weighted ensemble improves the result 1\% - to 3\% in development phase and 1\% - 2\% in evaluation phase - depending on the languages. For English this method performs the best in terms of ranking with $11^{th}$ rank while the worst for Moroccan Arabic with $21^{th}$ rank. On test Spearman correlation, English is the best securing 0.84 and Kinyarwanda is the worst with 0.37. Detailed results are shown in Table \ref{tab:combined_results1} of Appendix.

For Track B, embedding generated by language specific BERT based models provide the best result among the three combinations for all the languages. Then the average ensemble improves the result 0\% - to 3\% in development phase and 0\% - 2\% in evaluation phase - depending on the languages. For Kinyarwanda this method performs the best in terms of ranking with $1^{st}$ rank while the worst for English with $8^{th}$ rank. On test Spearman correlation, English is the best securing 0.77 and Punjabi is the worst with 0.02. Detailed result is shown in Table \ref{tab:combined_results2} of Appendix.

For Track C embedding generated by language specific (unrelated to target language) BERT based models provide the best result among the six combinations for all the languages. Then the average ensemble improves the result 0\% - to 2\% in both development and evaluation phases depending on the languages. For Punjabi this method performs the best in terms of ranking with $5^{th}$ rank while the worst for Hausa and Kinyarwanda with $12^{th}$ rank. On test Spearman correlation, Spanish is the best securing 0.56 and Punjabi is the worst with 0.02. Detailed result is shown in Table \ref{tab:combined_results3} of Appendix.

\section{Error Analysis}

\begin{figure*}
  \centering
  \includegraphics[width=0.96\textwidth]{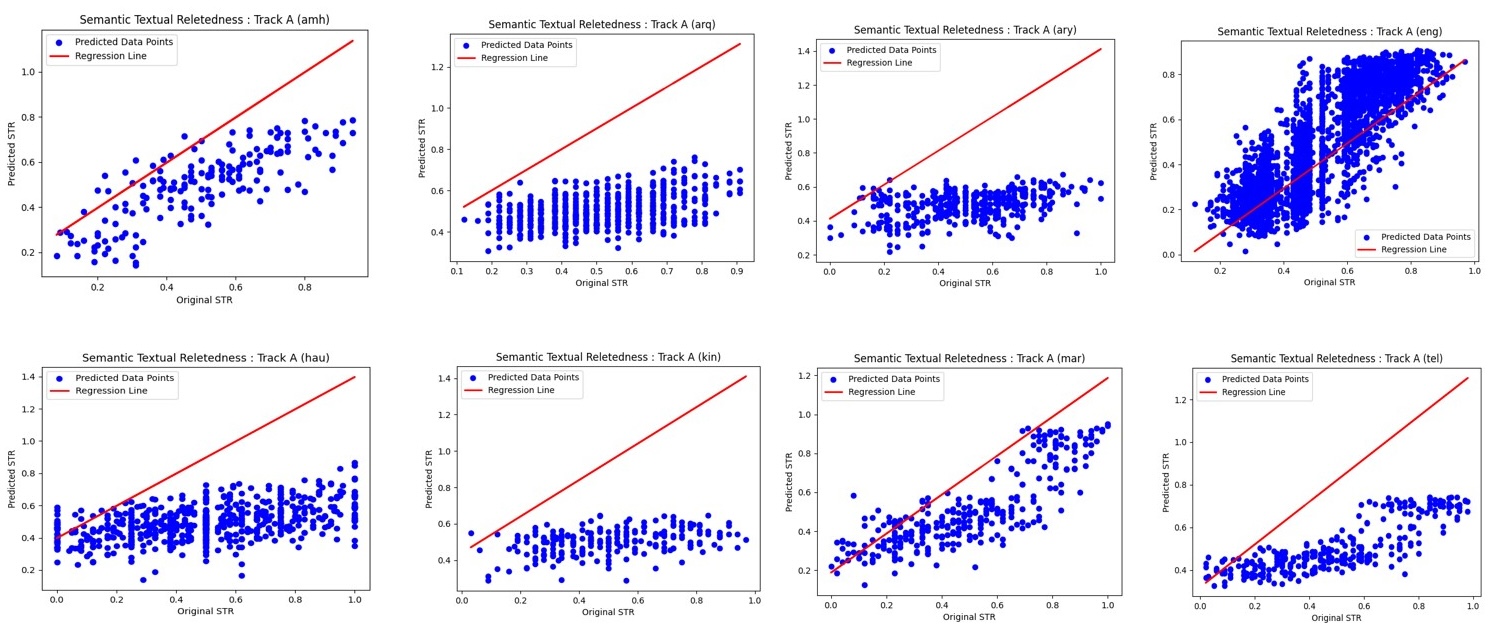}
  \caption{Track A (Comparison with gold semantic textual relatedness)}
  \label{fig:Track A}
\end{figure*}

\begin{figure*}
  \centering
  \includegraphics[width=0.96\textwidth]{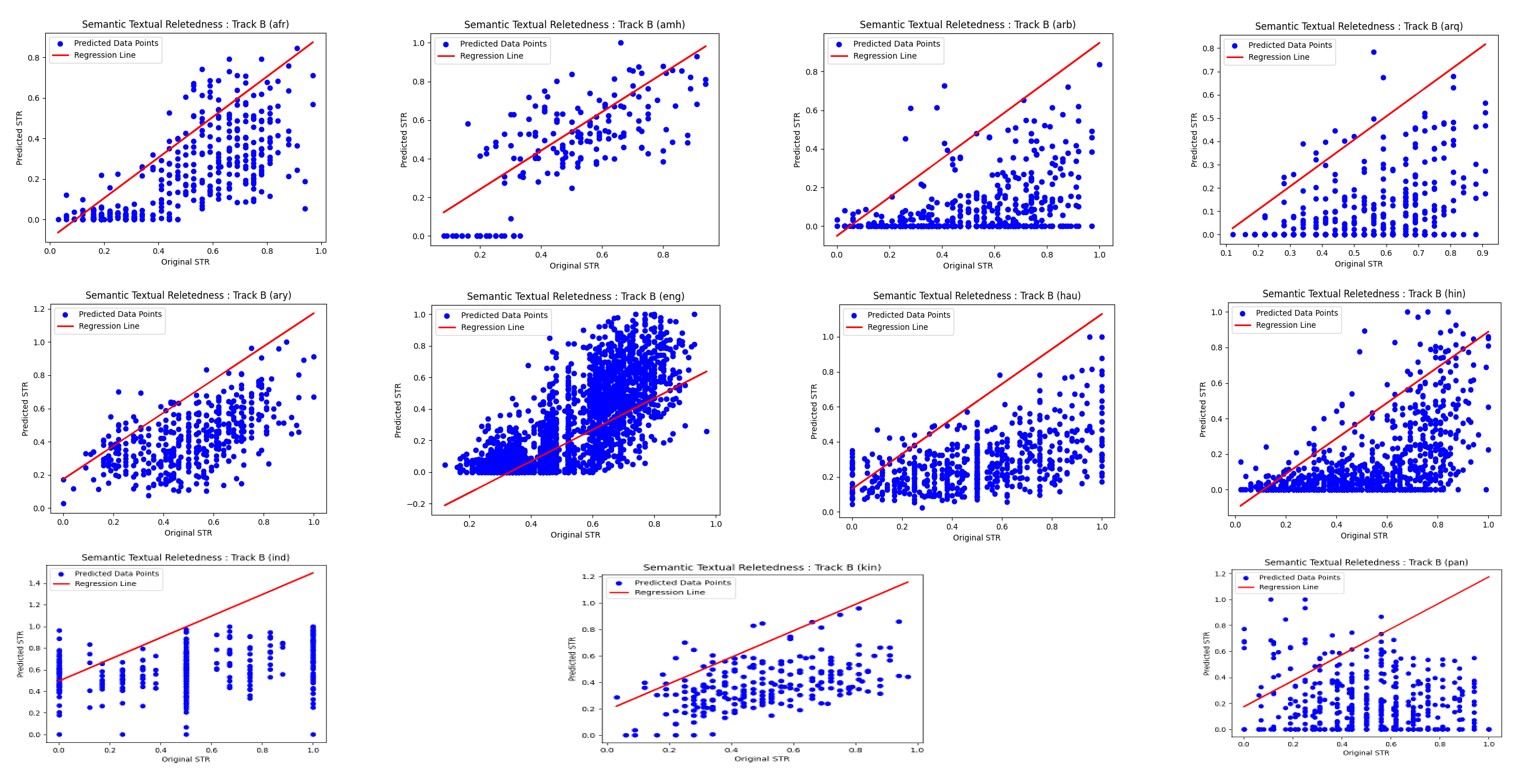}
  \caption{Track B (Comparison with gold semantic textual relatedness)}
  \label{fig:Track B}
\end{figure*}

\begin{figure*}
  \centering
  \includegraphics[width=0.96\textwidth]{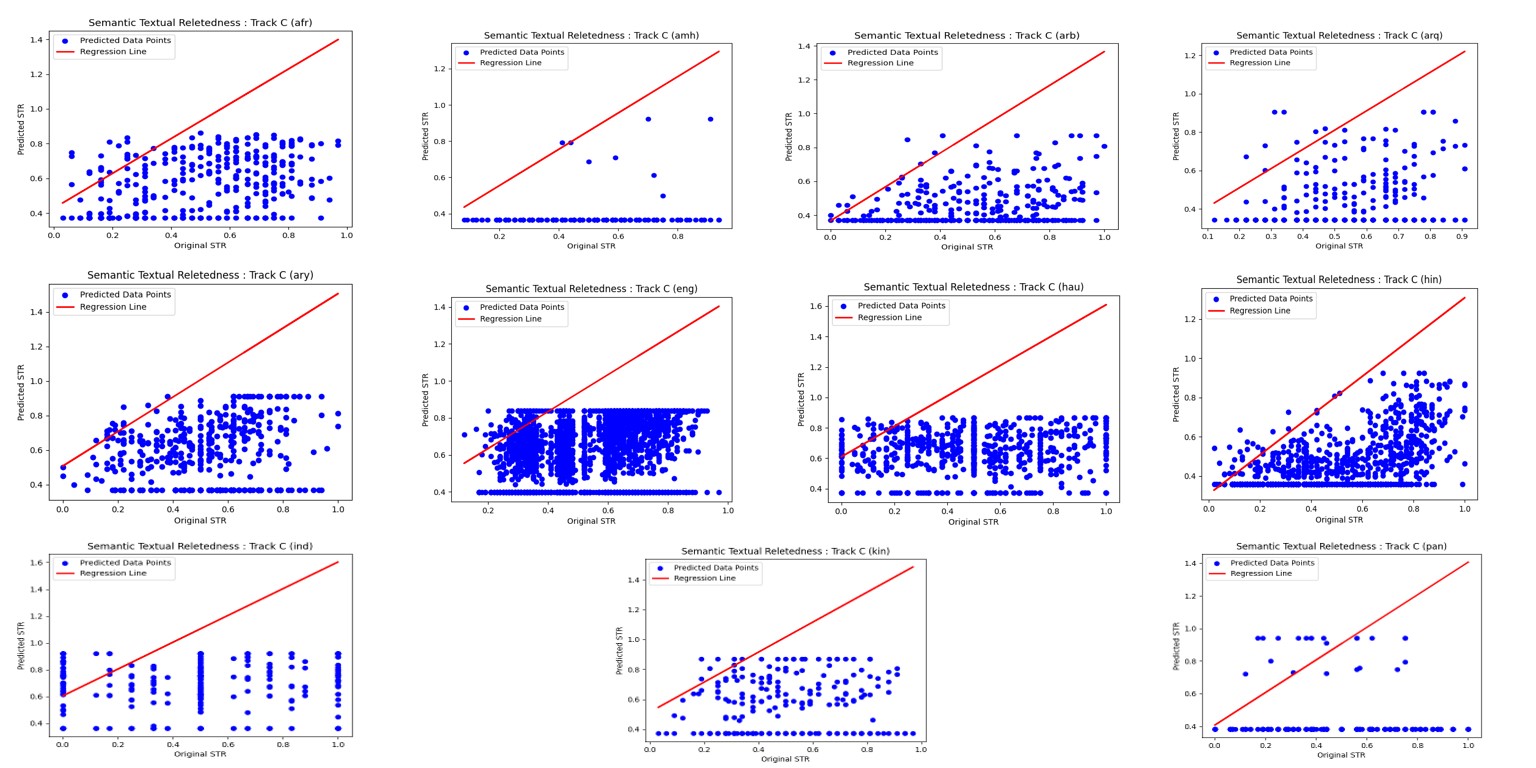}
  \caption{Track C (Comparison with gold semantic textual relatedness)}
  \label{fig:Track C}
\end{figure*}

For Track A, Algerian Arabic, Moroccan Arabic and Spanish test Spearman Correlation Coefficient decreases in the evaluation phase. This happens because the dev set was around 7.5\%-9\% and the test set is around 39\% - 46\% size of the train set.

For Track B, amount of dev data was only 20 for Afrikaans which is the reason of a very big difference between the result of development and evaluation phase. Algerian Arabic, Amharic, Modern Standard Arabic, Moroccan Arabic have a very small amount of dev data (less than 100) which is reason of decreased Spearman Correlation Coefficient in the evaluation phase. Hindi also faces the same issue but as it had more dev data the test Spearman Correlation Coefficient is only 4\% less than the development period.

For Track C, Algerian Arabic, Indonesian, Kinyarwanda, Modern Standard Arabic faced bigger drop of the Spearman Correlation Coefficient from the development phases. The main issue here is the BERT based models that doesn't know the target languages generate the embeddings that are not as good as what we observed in unsupervised track for the models with the knowledge of target language. Also the diversity of the train and test data make it more challenging to score better Spearman Correlation Coefficient. In addition, due to the unavailability of the text label, only the ensemble performance of Spanish language for all the tracks are shown.

Regarding the result of the Punjabi language in the both unsupervised and cross-lingual track, it was the most challenging language where the provide baseline was less than zero. Though our system achieves 0.02 Spearman Correlation Coefficient for for this language, the ranking is quite impressive which also proves the struggle of other teams to cope up with this language.

Moreover, ElasticNet and Linear Regression exhibit limitations as assumption of linearity may not align with the intricate and nonlinear relationships inherent in the textual data. The issue of dimensionality poses a challenge, especially when dealing with a large number of features. 
The difference between the gold and predicted semantic relatedness scores for the three tracks are shown in Figure \ref{fig:Track A}, Figure \ref{fig:Track B}, and Figure \ref{fig:Track C}.

\section{Conclusion}

We experimented with various methodologies on the dataset provided by the organizers, including statistical machine learning approaches, transformer-based models, language-specific BERTs, and sentence BERT. In the supervised task (Track A), with no restrictions on the model or data, we utilized the available training dataset. Conversely, the unsupervised task (Track B), lacking training data, presented challenges, leading us to use language-specific BERTs and statistical machine learning approaches. The cross-lingual track (Track C) imposed more stringent restrictions, requiring us to use training data from other languages in Track A, excluding the target language. In addition to statistical ML models, we integrated language-specific BERTs closely aligned with the geography and culture of the target language, as the use of LLMs was constrained due to unknown training data.

We show that our ensemble approach exhibited superior performance compared to individual model experiments. However, the task's inherent difficulty became evident in instances where relatively small datasets presented challenges for effective model learning. Semantic textual relatedness tasks face challenges like subjectivity, context dependency, and ambiguity due to multiple meanings and cultural differences. Limited data, domain specificity, short texts, and biases hinder accuracy. Ongoing research is crucial to address these limitations and improve model accuracy.


\section*{Acknowledgements}

We would like to thank the shared task organizers for providing participants with the dataset used in this paper.

\bibliography{anthology,custom}

\begin{thebibliography}{46}
\expandafter\ifx\csname natexlab\endcsname\relax\def\natexlab#1{#1}\fi

\bibitem[{Abdalla et~al.(2021)Abdalla, Vishnubhotla, and Mohammad}]{abdalla2021makes}
Mohamed Abdalla, Krishnapriya Vishnubhotla, and Saif~M Mohammad. 2021.
\newblock What makes sentences semantically related: A textual relatedness dataset and empirical study.
\newblock \emph{arXiv preprint arXiv:2110.04845}.

\bibitem[{Abdaoui et~al.(2021)Abdaoui, Berrimi, Oussalah, and Moussaoui}]{abdaoui2021dziribert}
Amine Abdaoui, Mohamed Berrimi, Mourad Oussalah, and Abdelouahab Moussaoui. 2021.
\newblock Dziribert: a pre-trained language model for the algerian dialect.
\newblock \emph{arXiv preprint arXiv:2109.12346}.

\bibitem[{Adelani et~al.(2022)Adelani, Neubig, Ruder, Rijhwani, Beukman, Palen-Michel, Lignos, Alabi, Muhammad, Nabende et~al.}]{adelani2022masakhaner}
David Adelani, Graham Neubig, Sebastian Ruder, Shruti Rijhwani, Michael Beukman, Chester Palen-Michel, Constantine Lignos, Jesujoba Alabi, Shamsuddeen Muhammad, Peter Nabende, et~al. 2022.
\newblock Masakhaner 2.0: Africa-centric transfer learning for named entity recognition.
\newblock In \emph{Proceedings of EMNLP}.

\bibitem[{Agirre et~al.(2015)Agirre, Banea, Cardie, Cer, Diab, Gonzalez-Agirre, Guo, Lopez-Gazpio, Maritxalar, Mihalcea et~al.}]{agirre2015semeval}
Eneko Agirre, Carmen Banea, Claire Cardie, Daniel Cer, Mona Diab, Aitor Gonzalez-Agirre, Weiwei Guo, Inigo Lopez-Gazpio, Montse Maritxalar, Rada Mihalcea, et~al. 2015.
\newblock Semeval-2015 task 2: Semantic textual similarity, english, spanish and pilot on interpretability.
\newblock In \emph{Proceedings of SemEval}.

\bibitem[{Agirre et~al.(2014)Agirre, Banea, Cardie, Cer, Diab, Gonzalez-Agirre, Guo, Mihalcea, Rigau, and Wiebe}]{agirre2014semeval}
Eneko Agirre, Carmen Banea, Claire Cardie, Daniel Cer, Mona Diab, Aitor Gonzalez-Agirre, Weiwei Guo, Rada Mihalcea, German Rigau, and Janyce Wiebe. 2014.
\newblock Semeval-2014 task 10: Multilingual semantic textual similarity.
\newblock In \emph{Proceedings of SemEval}.

\bibitem[{Agirre et~al.(2016)Agirre, Banea, Cer, Diab, Gonzalez~Agirre, Mihalcea, Rigau~Claramunt, and Wiebe}]{agirre2016semeval}
Eneko Agirre, Carmen Banea, Daniel Cer, Mona Diab, Aitor Gonzalez~Agirre, Rada Mihalcea, German Rigau~Claramunt, and Janyce Wiebe. 2016.
\newblock Semeval-2016 task 1: Semantic textual similarity, monolingual and cross-lingual evaluation.
\newblock In \emph{Proceedings of SemEval}.

\bibitem[{Agirre et~al.(2012)Agirre, Cer, Diab, and Gonzalez-Agirre}]{agirre2012semeval}
Eneko Agirre, Daniel Cer, Mona Diab, and Aitor Gonzalez-Agirre. 2012.
\newblock Semeval-2012 task 6: A pilot on semantic textual similarity.
\newblock In \emph{Proceedings of SemEval}.

\bibitem[{Agirre et~al.(2013)Agirre, Cer, Diab, Gonzalez-Agirre, and Guo}]{agirre2013sem}
Eneko Agirre, Daniel Cer, Mona Diab, Aitor Gonzalez-Agirre, and Weiwei Guo. 2013.
\newblock * sem 2013 shared task: Semantic textual similarity.
\newblock In \emph{Proceedings of * SEM}.

\bibitem[{Aizawa(2003)}]{aizawa2003information}
Akiko Aizawa. 2003.
\newblock An information-theoretic perspective of tf--idf measures.
\newblock \emph{Information Processing \& Management}.

\bibitem[{Ali et~al.(2009)Ali, GadAlla, and Abdelwahab}]{ali2009improving}
Ola~Mohammad Ali, Mahmoud GadAlla, and Mohammad~Said Abdelwahab. 2009.
\newblock Improving word sense disambiguation in machine translation using semantic relatedness and statistical measures of association.
\newblock In \emph{Proceedings of ICICIS}.

\bibitem[{Asaadi et~al.(2019)Asaadi, Mohammad, and Kiritchenko}]{asaadi2019big}
Shima Asaadi, Saif Mohammad, and Svetlana Kiritchenko. 2019.
\newblock Big bird: A large, fine-grained, bigram relatedness dataset for examining semantic composition.
\newblock In \emph{Proceedings of NAACL}.

\bibitem[{Baldissin et~al.(2022)Baldissin, Schlechtweg, and im~Walde}]{baldissin2022diawug}
Gioia Baldissin, Dominik Schlechtweg, and Sabine~Schulte im~Walde. 2022.
\newblock Diawug: A dataset for diatopic lexical semantic variation in spanish.
\newblock In \emph{Proceedings of LREC}.

\bibitem[{Bhattacharjee et~al.(2022)Bhattacharjee, Hasan, Ahmad, Mubasshir, Islam, Iqbal, Rahman, and Shahriyar}]{bhattacharjee2022banglabert}
Abhik Bhattacharjee, Tahmid Hasan, Wasi Ahmad, Kazi~Samin Mubasshir, Md~Saiful Islam, Anindya Iqbal, M.~Sohel Rahman, and Rifat Shahriyar. 2022.
\newblock {B}angla{BERT}: Language model pretraining and benchmarks for low-resource language understanding evaluation in {B}angla.
\newblock In \emph{Findings of NAACL}.

\bibitem[{Cañete et~al.(2020)Cañete, Chaperon, Fuentes, Ho, Kang, and Pérez}]{CaneteCFP2020}
José Cañete, Gabriel Chaperon, Rodrigo Fuentes, Jou-Hui Ho, Hojin Kang, and Jorge Pérez. 2020.
\newblock Spanish pre-trained bert model and evaluation data.
\newblock In \emph{Proccedings of PML4DC (ICLR)}.

\bibitem[{Church and Hanks(1990)}]{church1990word}
Kenneth Church and Patrick Hanks. 1990.
\newblock Word association norms, mutual information, and lexicography.
\newblock \emph{Computational linguistics}.

\bibitem[{Cruz and Cheng(2021)}]{cruz2021improving}
Jan Christian~Blaise Cruz and Charibeth Cheng. 2021.
\newblock Improving large-scale language models and resources for filipino.
\newblock \emph{arXiv preprint arXiv:2111.06053}.

\bibitem[{David(1963)}]{david1963method}
Herbert~Aron David. 1963.
\newblock \emph{The method of paired comparisons}.
\newblock London.

\bibitem[{Dolan and Brockett(2005)}]{dolan2005automatically}
Bill Dolan and Chris Brockett. 2005.
\newblock Automatically constructing a corpus of sentential paraphrases.
\newblock In \emph{Procedings of IWP)}.

\bibitem[{Feng et~al.(2020)Feng, Yang, Cer, Arivazhagan, and Wang}]{feng2020language}
Fangxiaoyu Feng, Yinfei Yang, Daniel Cer, Naveen Arivazhagan, and Wei Wang. 2020.
\newblock Language-agnostic bert sentence embedding.
\newblock \emph{arXiv preprint arXiv:2007.01852}.

\bibitem[{Gaanoun et~al.(2024)Gaanoun, Naira, Allak, and Benelallam}]{gaanoun2024darijabert}
Kamel Gaanoun, Abdou~Mohamed Naira, Anass Allak, and Imade Benelallam. 2024.
\newblock Darijabert: a step forward in nlp for the written moroccan dialect.
\newblock \emph{International Journal of Data Science and Analytics}.

\bibitem[{Goswami et~al.(2024)Goswami, Puspo, Raihan, and Emran}]{goswami2024masontigers}
Dhiman Goswami, Sadiya Sayara~Chowdhury Puspo, Md~Nishat Raihan, and Al~Nahian~Bin Emran. 2024.
\newblock Masontigers@ {LT-EDI}-2024: An ensemble approach towards detecting homophobia and transphobia in social media comments.
\newblock \emph{arXiv preprint arXiv:2401.14681}.

\bibitem[{Gro{\ss}(2003)}]{gross2003linear}
J{\"u}rgen Gro{\ss}. 2003.
\newblock \emph{Linear regression}.
\newblock Springer Science \& Business Media.

\bibitem[{Halliday and Hasan(2014)}]{halliday2014cohesion}
Michael Alexander~Kirkwood Halliday and Ruqaiya Hasan. 2014.
\newblock \emph{Cohesion in english}.
\newblock Routledge.

\bibitem[{Hasan and Halliday(1976)}]{hasan1976cohesion}
Ruqaiya Hasan and Michael~AK Halliday. 1976.
\newblock Cohesion in english.
\newblock \emph{London, 1976; Martin JR}.

\bibitem[{Joshi(2022{\natexlab{a}})}]{joshi2022l3cubehind}
Raviraj Joshi. 2022{\natexlab{a}}.
\newblock L3cube-hindbert and devbert: Pre-trained bert transformer models for devanagari based hindi and marathi languages.
\newblock \emph{arXiv preprint arXiv:2211.11418}.

\bibitem[{Joshi(2022{\natexlab{b}})}]{joshi2022}
Raviraj Joshi. 2022{\natexlab{b}}.
\newblock {L}3{C}ube-{M}aha{C}orpus and {M}aha{BERT}: {M}arathi monolingual corpus, {M}arathi {BERT} language models, and resources.
\newblock In \emph{Proceedings of WILDRE}.

\bibitem[{Kiritchenko and Mohammad(2016)}]{kiritchenko2016capturing}
Svetlana Kiritchenko and Saif~M Mohammad. 2016.
\newblock Capturing reliable fine-grained sentiment associations by crowdsourcing and best--worst scaling.
\newblock In \emph{Proceedings of HLT (NAACL)}.

\bibitem[{Kiritchenko and Mohammad(2017)}]{kiritchenko2017best}
Svetlana Kiritchenko and Saif~M Mohammad. 2017.
\newblock Best-worst scaling more reliable than rating scales: A case study on sentiment intensity annotation.
\newblock \emph{arXiv preprint arXiv:1712.01765}.

\bibitem[{Koto et~al.(2020)Koto, Rahimi, Lau, and Baldwin}]{koto2020indolem}
Fajri Koto, Afshin Rahimi, Jey~Han Lau, and Timothy Baldwin. 2020.
\newblock Indolem and indobert: A benchmark dataset and pre-trained language model for indonesian nlp.
\newblock In \emph{Proceedings of COLING}.

\bibitem[{Li et~al.(2006)Li, McLean, Bandar, O'shea, and Crockett}]{li2006sentence}
Yuhua Li, David McLean, Zuhair~A Bandar, James~D O'shea, and Keeley Crockett. 2006.
\newblock Sentence similarity based on semantic nets and corpus statistics.
\newblock \emph{IEEE transactions on knowledge and data engineering}.

\bibitem[{Liu et~al.(2019)Liu, Ott, Goyal, Du, Joshi, Chen, Levy, Lewis, Zettlemoyer, and Stoyanov}]{liu2019roberta}
Yinhan Liu, Myle Ott, Naman Goyal, Jingfei Du, Mandar Joshi, Danqi Chen, Omer Levy, Mike Lewis, Luke Zettlemoyer, and Veselin Stoyanov. 2019.
\newblock Roberta: {A} robustly optimized {BERT} pretraining approach.
\newblock \emph{CoRR}, abs/1907.11692.

\bibitem[{Louviere and Woodworth(1991)}]{louviere1991best}
Jordan~J Louviere and George~G Woodworth. 1991.
\newblock Best-worst scaling: A model for the largest difference judgments.
\newblock Technical report, Working paper.

\bibitem[{Morris and Hirst(1991)}]{morris1991lexical}
Jane Morris and Graeme Hirst. 1991.
\newblock Lexical cohesion computed by thesaural relations as an indicator of the structure of text.
\newblock \emph{Computational linguistics}.

\bibitem[{Myers and Sirois(2004)}]{myers2004spearman}
Leann Myers and Maria~J Sirois. 2004.
\newblock Spearman correlation coefficients, differences between.
\newblock \emph{Encyclopedia of statistical sciences}.

\bibitem[{Nzeyimana and Niyongabo~Rubungo(2022)}]{nzeyimana2022kinyabert}
Antoine Nzeyimana and Andre Niyongabo~Rubungo. 2022.
\newblock {K}inya{BERT}: a morphology-aware {K}inyarwanda language model.
\newblock In \emph{Proceedings of ACL}.

\bibitem[{Ousidhoum et~al.(2024{\natexlab{a}})Ousidhoum, Muhammad, Abdalla, Abdulmumin, Ahmad, Ahuja, Aji, Araujo, Ayele, Baswani, Beloucif, Biemann, Bourhim, Kock, Dekebo, Hourrane, Kanumolu, Madasu, Rutunda, Shrivastava, Solorio, Surange, Tilaye, Vishnubhotla, Winata, Yimam, and Mohammad}]{ousidhoum2024semrel2024}
Nedjma Ousidhoum, Shamsuddeen~Hassan Muhammad, Mohamed Abdalla, Idris Abdulmumin, Ibrahim~Said Ahmad, Sanchit Ahuja, Alham~Fikri Aji, Vladimir Araujo, Abinew~Ali Ayele, Pavan Baswani, Meriem Beloucif, Chris Biemann, Sofia Bourhim, Christine~De Kock, Genet~Shanko Dekebo, Oumaima Hourrane, Gopichand Kanumolu, Lokesh Madasu, Samuel Rutunda, Manish Shrivastava, Thamar Solorio, Nirmal Surange, Hailegnaw~Getaneh Tilaye, Krishnapriya Vishnubhotla, Genta Winata, Seid~Muhie Yimam, and Saif~M. Mohammad. 2024{\natexlab{a}}.
\newblock \href {http://arxiv.org/abs/2402.08638} {Semrel2024: A collection of semantic textual relatedness datasets for 14 languages}.

\bibitem[{Ousidhoum et~al.(2024{\natexlab{b}})Ousidhoum, Muhammad, Abdalla, Abdulmumin, Ahmad, Ahuja, Aji, Araujo, Beloucif, De~Kock, Hourrane, Shrivastava, Solorio, Surange, Vishnubhotla, Yimam, and Mohammad}]{semrel2024task}
Nedjma Ousidhoum, Shamsuddeen~Hassan Muhammad, Mohamed Abdalla, Idris Abdulmumin, Ibrahim~Said Ahmad, Sanchit Ahuja, Alham~Fikri Aji, Vladimir Araujo, Meriem Beloucif, Christine De~Kock, Oumaima Hourrane, Manish Shrivastava, Thamar Solorio, Nirmal Surange, Krishnapriya Vishnubhotla, Seid~Muhie Yimam, and Saif~M. Mohammad. 2024{\natexlab{b}}.
\newblock {S}em{E}val-2024 task 1: Semantic textual relatedness for african and asian languages.
\newblock In \emph{Proceedings of the 18th International Workshop on Semantic Evaluation (SemEval-2024)}. Association for Computational Linguistics.

\bibitem[{Park et~al.(2014)Park, Shim, and Lee}]{park2014isoft}
Seonyeong Park, Hyosup Shim, and Gary~Geunbae Lee. 2014.
\newblock Isoft at qald-4: Semantic similarity-based question answering system over linked data.
\newblock In \emph{CLEF (Working Notes)}.

\bibitem[{Piao et~al.(2016)Piao, Ara, and Breslin}]{piao2016computing}
Guangyuan Piao, Safina~Showkat Ara, and John~G Breslin. 2016.
\newblock Computing the semantic similarity of resources in dbpedia for recommendation purposes.
\newblock In \emph{Proceedings of JIST}.

\bibitem[{Rahman and Borah(2023)}]{rahman2023query}
Nazreena Rahman and Bhogeswar Borah. 2023.
\newblock Query-based extractive text summarization using sense-oriented semantic relatedness measure.
\newblock \emph{Arabian Journal for Science and Engineering}.

\bibitem[{Rahutomo et~al.(2012)Rahutomo, Kitasuka, and Aritsugi}]{rahutomo2012semantic}
Faisal Rahutomo, Teruaki Kitasuka, and Masayoshi Aritsugi. 2012.
\newblock Semantic cosine similarity.
\newblock In \emph{Proceedings of ICAST}.

\bibitem[{Rohrbach et~al.(2010)Rohrbach, Stark, Szarvas, Gurevych, and Schiele}]{rohrbach2010helps}
Marcus Rohrbach, Michael Stark, Gy{\"o}rgy Szarvas, Iryna Gurevych, and Bernt Schiele. 2010.
\newblock What helps where--and why? semantic relatedness for knowledge transfer.
\newblock In \emph{Proceedings of CVPR}.

\bibitem[{Safaya et~al.(2020)Safaya, Abdullatif, and Yuret}]{safaya2020kuisail}
Ali Safaya, Moutasem Abdullatif, and Deniz Yuret. 2020.
\newblock {KUISAIL} at {S}em{E}val-2020 task 12: {BERT}-{CNN} for offensive speech identification in social media.
\newblock In \emph{Proceedings of SemEval}.

\bibitem[{Thurstone(1994)}]{thurstone1994law}
Louis~L Thurstone. 1994.
\newblock A law of comparative judgment.
\newblock \emph{Psychological review}.

\bibitem[{Yimam et~al.(2021)Yimam, Ayele, Venkatesh, Gashaw, and Biemann}]{yimam2021introducing}
Seid~Muhie Yimam, Abinew~Ali Ayele, Gopalakrishnan Venkatesh, Ibrahim Gashaw, and Chris Biemann. 2021.
\newblock Introducing various semantic models for amharic: Experimentation and evaluation with multiple tasks and datasets.
\newblock \emph{Future Internet}, 13.

\bibitem[{Zou and Hastie(2005)}]{zou2005regularization}
Hui Zou and Trevor Hastie. 2005.
\newblock Regularization and variable selection via the elastic net.
\newblock \emph{Journal of the Royal Statistical Society Series B: Statistical Methodology}.

\end{thebibliography}

\appendix

\section{Appendix}

\vspace{.01cm}

\begin{table*}[!h]
\centering
\small
\setlength{\tabcolsep}{14pt}
\begin{tabular}{lcc|lcc}
\hline
\multicolumn{3}{|c|}{\textbf{Algerian Arabic (arq) - (Rank 19)}} & \multicolumn{3}{c|}{\textbf{Marathi (mar) - (Rank 19)}} \\
\hline
Models & Dev SC & Test SC & Models & Dev SC & Test SC \\
\hline
TF-IDF + EN & 0.43 & 0.33 & TF-IDF + EN & 0.65 & 0.76 \\
PPMI + EN & 0.44 & 0.34 & PPMI + EN & 0.67 & 0.77 \\
DziriBERT + EN & 0.44 & 0.35 & MarathiBERT + EN & 0.68 & 0.80 \\
LaBSE + EN & 0.46 & 0.36 & LaBSE + EN & 0.68 & 0.79\\
\hline
TF-IDF + LR & 0.45 & 0.34 & TF-IDF + LR & 0.67 & 0.79 \\
PPMI + LR & 0.46 & 0.37 & PPMI + LR & 0.67 & 0.80 \\
DziriBERT + LR & 0.48 & 0.37 & MarathiBERT + LR & 0.69 & 0.81\\
LaBSE + LR & 0.48 & 0.38 & LaBSE + LR & 0.69 & 0.81\\
\hline
Wt. (Dev. SC) Ensemble  & 0.49 & 0.40 & Wt. (Dev. SC) Ensemble  & 0.70 & 0.82 \\
\hline
\multicolumn{3}{|c|}{\textbf{Amharic (amh) - (Rank 12)}} & \multicolumn{3}{c|}{\textbf{Moroccan Arabic (ary) - (Rank 21)}} \\
\hline
Models & Dev SC & Test SC & Models & Dev SC & Test SC \\
\hline
TF-IDF + EN & 0.67 & 0.74 & TF-IDF + EN & 0.41 & 0.30 \\
PPMI + EN & 0.68 & 0.76 & PPMI + EN & 0.43 & 0.33 \\
AmRoBERTa + EN & 0.68 & 0.76 & DarijaBERT + EN & 0.44 & 0.34 \\
LaBSE + EN & 0.68 & 0.77 & LaBSE + EN & 0.45 & 0.34\\
\hline
TF-IDF + LR & 0.67 & 0.75 & TF-IDF + LR & 0.44 & 0.34 \\
PPMI + LR & 0.69 & 0.77 & PPMI + LR & 0.45 & 0.35\\
AmRoBERTa + LR & 0.70 & 0.78 & DarijaBERT + LR & 0.46 & 0.36\\
LaBSE + LR & 0.70 & 0.78 & LaBSE + LR & 0.46 & 0.36\\
\hline
Wt. (Dev. SC) Ensemble  & 0.71 & 0.79 & Wt. (Dev. SC) Ensemble  & 0.48 & 0.38 \\
\hline
\multicolumn{3}{|c|}{\textbf{English (eng) - (Rank 11)}} & \multicolumn{3}{c|}{\textbf{Spanish (esp) - (Rank 19)}} \\
\hline
Models & Dev SC & Test SC & Models & Dev SC & Test SC \\
\hline
TF-IDF + EN & 0.76 & 0.78 & TF-IDF + EN & 0.58 &  \\
PPMI + EN & 0.78 & 0.80 & PPMI + EN & 0.58 &  \\
RoBERTa + EN & 0.79 & 0.82 & SpanishBERT + EN & 0.61 & \\
LaBSE + EN & 0.80 & 0.82 & LaBSE + EN & 0.63 & \\
\hline
TF-IDF + LR & 0.78 & 0.81  & TF-IDF + LR & 0.62 &  \\
PPMI + LR & 0.79 & 0.82 & PPMI + LR & 0.62 &  \\
RoBERTa + LR & 0.80 & 0.83 & SpanishBERT + LR & 0.63 & \\
LaBSE + LR & 0.80 & 0.83 & LaBSE + LR & 0.63 & \\
\hline
Wt. (Dev. SC) Ensemble  & 0.81 & 0.84 & Wt. (Dev. SC) Ensemble  & 0.66 & 0.65\\
\hline
\multicolumn{3}{|c|}{\textbf{Hausa (hau) - (Rank 18)}} & \multicolumn{3}{c|}{\textbf{Telugu (tel) - (Rank 13)}} \\
\hline
Models & Dev SC & Test SC & Models & Dev SC & Test SC \\
\hline
TF-IDF + EN & 0.31 & 0.42 & TF-IDF + EN & 0.71 & 0.72 \\
PPMI + EN & 0.33 & 0.45 & PPMI + EN & 0.74 & 0.76 \\
HauRoBERTa + EN & 0.34 & 0.46 & TeluguBERT + EN & 0.75 & 0.77\\
LaBSE + EN & 0.34 & 0.46  & LaBSE + EN & 0.75 & 0.77\\
\hline
TF-IDF + LR & 0.32 & 0.41 & TF-IDF + LR & 0.74 &  0.75\\
PPMI + LR & 0.33 & 0.45 & PPMI + LR & 0.74 & 0.76 \\
HauRoBERTa + LR & 0.35 & 0.46 & TeluguBERT + LR & 0.75 & 0.77\\
LaBSE + LR & 0.35 & 0.47 & LaBSE + LR & 0.76 & 0.78\\
\hline
Wt. (Dev. SC) Ensemble  & 0.36 & 0.48 & Wt. (Dev. SC) Ensemble  & 0.78 & 0.80\\
\hline
\multicolumn{3}{|c|}{\textbf{Kinyarwanda (kin) - (Rank 18)}} \\
\cline{1-3}
Models & Dev SC & Test SC \\
\cline{1-3}
TF-IDF + EN & 0.23 &  0.31  \\
PPMI + EN & 0.25 &  0.33  \\
KinyaBERT + EN & 0.25 & 0.34 \\
LaBSE + EN & 0.25 & 0.34\\
\cline{1-3}
TF-IDF + LR & 0.25 & 0.33 \\
PPMI + LR & 0.25 & 0.33 \\
KinyaBERT + LR & 0.25 & 0.34\\
LaBSE + LR & 0.27 & 0.35\\
\cline{1-3}
Wt. (Dev. SC) Ensemble & 0.28 & 0.37\\
\cline{1-3}
\end{tabular}
\caption{Results of Track A (Supervised) (EN : ElasticNet, LR : Linear Regression, SC : Spearman correlation)}
\label{tab:combined_results1}
\end{table*}


\begin{table*}[!h]
\centering
\small
\setlength{\tabcolsep}{17pt}
\begin{tabular}{lcc|lcc}
\hline
\multicolumn{3}{|c|}{\textbf{Afrikaans (afr) - (Rank 4)}} & \multicolumn{3}{c|}{\textbf{Indonesian (ind) - (Rank 6)}} \\
\hline
Models & Dev SC & Test SC & Models & Dev SC & Test SC \\
\hline
TF-IDF & 0.01 & 0.73 & TF-IDF & 0.31 & 0.33 \\
PPMI & 0.02 & 0.73 & PPMI & 0.32 & 0.35 \\
AfricanBERTa & 0.02 & 0.74 & IndoBERT & 0.33 & 0.36\\
\hline
Ensemble  & 0.02 & 0.76 & Ensemble  &  0.35 & 0.38\\
\hline
\multicolumn{3}{|c|}{\textbf{Algerian Arabic (arq) - (Rank 3)}} & \multicolumn{3}{c|}{\textbf{Kinyarwanda 
 (kin) - (Rank 1)}} \\
\hline
Models & Dev SC & Test SC & Models & Dev SC & Test SC \\
\hline
TF-IDF & 0.45 & 0.36 & TF-IDF & 0.13 & 0.42 \\
PPMI & 0.48 & 0.38 & PPMI & 0.14 & 0.44 \\
DziriBERT & 0.49 & 0.40 & KinyaBERT & 0.14 & 0.45\\
\hline
Ensemble  & 0.52 & 0.42 & Ensemble  &  0.15 & 0.46\\
\hline
\multicolumn{3}{|c|}{\textbf{Amharic (amh) - (Rank 3)}} & \multicolumn{3}{c|}{\textbf{Modern Standard Arabic (arb) - (Rank 4)}} \\
\hline
Models & Dev SC & Test SC & Models & Dev SC & Test SC \\
\hline
TF-IDF & 0.61 & 0.61 & TF-IDF & 0.40 & 0.37 \\
PPMI & 0.63 & 0.63 & PPMI & 0.41 & 0.38 \\
AmRoBERTa & 0.66 & 0.65 & ArabicBERT & 0.41 & 0.39\\
\hline
Ensemble  & 0.67 & 0.66 & Ensemble  &  0.42 & 0.40\\
\hline
\multicolumn{3}{|c|}{\textbf{English (eng) - (Rank 8)}} & \multicolumn{3}{c|}{\textbf{Moroccan Arabic (ary) - (Rank 2)}} \\
\hline
Models & Dev SC & Test SC & Models & Dev SC & Test SC \\
\hline
TF-IDF & 0.63 & 0.72 & TF-IDF & 0.61 & 0.51 \\
PPMI & 0.65 & 0.74 & PPMI & 0.63 & 0.54 \\
RoBERTa & 0.66 & 0.75 & DarijaBERT & 0.63 & 0.55\\
\hline
Ensemble  & 0.68 & 0.77 & Ensemble  &  0.65 & 0.56\\
\hline
\multicolumn{3}{|c|}{\textbf{Hausa (hau) - (Rank 2)}} & \multicolumn{3}{c|}{\textbf{Punjabi (pan) - (Rank 2)}} \\
\hline
Models & Dev SC & Test SC & Models & Dev SC & Test SC \\
\hline
TF-IDF & 0.42 & 0.45 & TF-IDF & 0.03 & 0.01 \\
PPMI & 0.45 & 0.47 & PPMI & 0.03 & 0.01 \\
HauRoBERTa & 0.46 & 0.48 & PunjabiBERT & 0.04 & 0.02\\
\hline
Ensemble  & 0.47 & 0.50 & Ensemble  &  0.04 & 0.02\\
\hline
\multicolumn{3}{|c|}{\textbf{Hindi (hin) - (Rank 7)}} & \multicolumn{3}{c|}{\textbf{Spanish (esp) - (Rank 4)}} \\
\hline
Models & Dev SC & Test SC & Models & Dev SC & Test SC \\
\hline
TF-IDF & 0.58 & 0.53 & TF-IDF & 0.57 &  \\
PPMI & 0.58 & 0.54 & PPMI & 0.58 &  \\
HindiBERT & 0.60 & 0.56 & SpanishBERT & 0.59 & \\
\hline
Ensemble  & 0.61 & 0.57 & Ensemble  &  0.60 & 0.66\\
\hline
\end{tabular}%

\caption{Results for Track B (Unsupervised) (EN : ElasticNet, LR : Linear Regression, SC : Spearman Correlation)}
\label{tab:combined_results2}
\end{table*}


\begin{table*}[!h]
\centering
\small
\setlength{\tabcolsep}{16pt}
\begin{tabular}{lcc|lcc}
\hline
\multicolumn{3}{|c|}{\textbf{Afrikaans (afr) - (Rank 11)}} & \multicolumn{3}{c|}{\textbf{Indonesian (ind) - (Rank 11)}} \\
\hline
Models & Dev SC & Test SC & Models & Dev SC & Test SC \\
\hline
TF-IDF + EN & 0.07 & 0.33 & TF-IDF + EN & 0.24 &  0.10\\
PPMI + EN & 0.08 & 0.35 & PPMI + EN & 0.25 & 0.11 \\
ArabicBERT + EN & 0.09 & 0.35 & RoBERTa-tagalog  + EN & 0.27 & 0.12\\
\hline
TF-IDF + LR & 0.08 & 0.34 & TF-IDF + LR & 0.26 & 0.11 \\
PPMI + LR & 0.10 & 0.36 & PPMI + LR & 0.27 & 0.12 \\
ArabicBERT + LR & 0.10 & 0.37 & RoBERTa-tagalog  + LR & 0.27 & 0.13\\
\hline
Ensemble  & 0.11 & 0.38 & Ensemble  &  0.29 & 0.13\\
\hline
\multicolumn{3}{|c|}{\textbf{Algerian Arabic (arq) - (Rank 9)}} & \multicolumn{3}{c|}{\textbf{Kinyarwanda 
 (kin) - (Rank 12)}} \\
\hline
Models & Dev SC & Test SC & Models & Dev SC & Test SC \\
\hline
TF-IDF + EN & 0.25 & 0.17 & TF-IDF + EN & 0.22 &  0.03\\
PPMI + EN & 0.27 & 0.19 & PPMI + EN & 0.23 &  0.04\\
AfricanBERTa + EN & 0.27 & 0.20 & ArabicBERT  + EN & 0.26 & 0.06\\
\hline
TF-IDF + LR & 0.27 & 0.19 & TF-IDF + LR & 0.24 &  0.05\\
PPMI + LR & 0.28 & 0.21 & PPMI + LR & 0.25 &  0.06\\
AfricanBERTa + LR & 0.29 & 0.21 & ArabicBERT  + LR & 0.26 & 0.07\\
\hline
Ensemble  & 0.30 & 0.22 & Ensemble  &  0.28 & 0.08\\
\hline
\multicolumn{3}{|c|}{\textbf{Amharic (amh) - (Rank 9)}} & \multicolumn{3}{c|}{\textbf{Modern Standard Arabic (arb) - (Rank 8)}} \\
\hline
Models & Dev SC & Test SC & Models & Dev SC & Test SC \\
\hline
TF-IDF + EN & 0.06 & 0.08 & TF-IDF + EN & 0.21 & 0.15 \\
PPMI + EN & 0.09 & 0.09 & PPMI + EN & 0.24 & 0.18 \\
ArabicBERT + EN & 0.09 & 0.10 & AfricanBERTa  + EN & 0.25 & 0.18\\
\hline
TF-IDF + LR & 0.09 & 0.10 & TF-IDF + LR & 0.22 &  0.16\\
PPMI + LR & 0.10 & 0.11 & PPMI + LR & 0.25 & 0.18 \\
ArabicBERT + LR & 0.10  & 0.12 & AfricanBERTa  + LR & 0.26 & 0.19\\
\hline
Ensemble  & 0.11 & 0.13 & Ensemble  &  0.27 & 0.21\\
\hline
\multicolumn{3}{|c|}{\textbf{English (eng) - (Rank 9)}} & \multicolumn{3}{c|}{\textbf{Moroccan Arabic (ary) - (Rank 10)}} \\
\hline
Models & Dev SC & Test SC & Models & Dev SC & Test SC \\
\hline
TF-IDF + EN & 0.25 & 0.26 & TF-IDF + EN & 0.09 & 0.14 \\
PPMI + EN & 0.26 & 0.27 & PPMI + EN & 0.12 & 0.17 \\
SpanishBERT + EN & 0.28 & 0.29 & AfricanBERTa  + EN & 0.12 & 0.18\\
\hline
TF-IDF + LR & 0.26 & 0.28 & TF-IDF + LR & 0.10 & 0.15 \\
PPMI + LR & 0.27 & 0.28 & PPMI + LR & 0.13 &  0.17\\
SpanishBERT + LR & 0.28 & 0.30 & AfricanBERTa  + LR & 0.14 & 0.19\\
\hline
Ensemble  & 0.29 & 0.31 & Ensemble  &  0.15 & 0.20\\
\hline
\multicolumn{3}{|c|}{\textbf{Hausa (hau) - (Rank 12)}} & \multicolumn{3}{c|}{\textbf{Punjabi (pan) - (Rank 5)}} \\
\hline
Models & Dev SC & Test SC & Models & Dev SC & Test SC \\
\hline
TF-IDF + EN & 0.08 & 0.06 & TF-IDF + EN & 0.01 & 0.01 \\
PPMI + EN & 0.09 & 0.07 & PPMI + EN & 0.02 &  0.01\\
ArabicBERT + EN & 0.11 & 0.08 & HindiBERT  + EN & 0.03 & 0.02\\
\hline
TF-IDF + LR & 0.09 & 0.07 & TF-IDF + LR & 0.02 &  0.01\\
PPMI + LR & 0.10 & 0.07 & PPMI + LR & 0.03 & 0.02 \\
ArabicBERT + LR & 0.11 & 0.09 & HindiBERT  + LR & 0.04 & 0.02\\
\hline
Ensemble  & 0.12 & 0.10 & Ensemble  &  0.04 & 0.02 \\
\hline
\multicolumn{3}{|c|}{\textbf{Hindi (hin) - (Rank 9)}} & \multicolumn{3}{c|}{\textbf{Spanish (esp) - (Rank 10)}} \\
\hline
Models & Dev SC & Test SC & Models & Dev SC & Test SC \\
\hline
TF-IDF + EN & 0.48 & 0.43 & TF-IDF + EN & 0.39 &  \\
PPMI + EN & 0.51 & 0.47 & PPMI + EN & 0.40 &  \\
BanglaBERT + EN & 0.53 & 0.49 & roBERTa  + EN &  0.43 & \\
\hline
TF-IDF + LR & 0.50 & 0.46 & TF-IDF + LR & 0.41 &  \\
PPMI + LR & 0.52 & 0.48 & PPMI + LR & 0.42 &  \\
BanglaBERT + LR & 0.53 & 0.50 & roBERTa  + LR &  0.44 & \\
\hline
Ensemble  & 0.55 & 0.51 & Ensemble  &  0.45 & 0.56\\
\hline
\end{tabular}%
\caption{Results for Track C (Cross-lingual) (EN : ElasticNet, LR : Linear Regression, SC : Spearman Correlation)}
\label{tab:combined_results3}
\end{table*}

\end{document}